%% file: main.tex
\documentclass[]{lbw}
\definecolor{metablue}{HTML}{1772B4}
\definecolor{tablerowcolor}{RGB}{235,245,252}
\definecolor{tablerowcolor1}{RGB}{235,248,235}
\definecolor{mygreen}{RGB}{84,130,53}
\setabstractbrandtext{}
\setabstractbrandcolor{metablue}
\setabstractbrandlogo{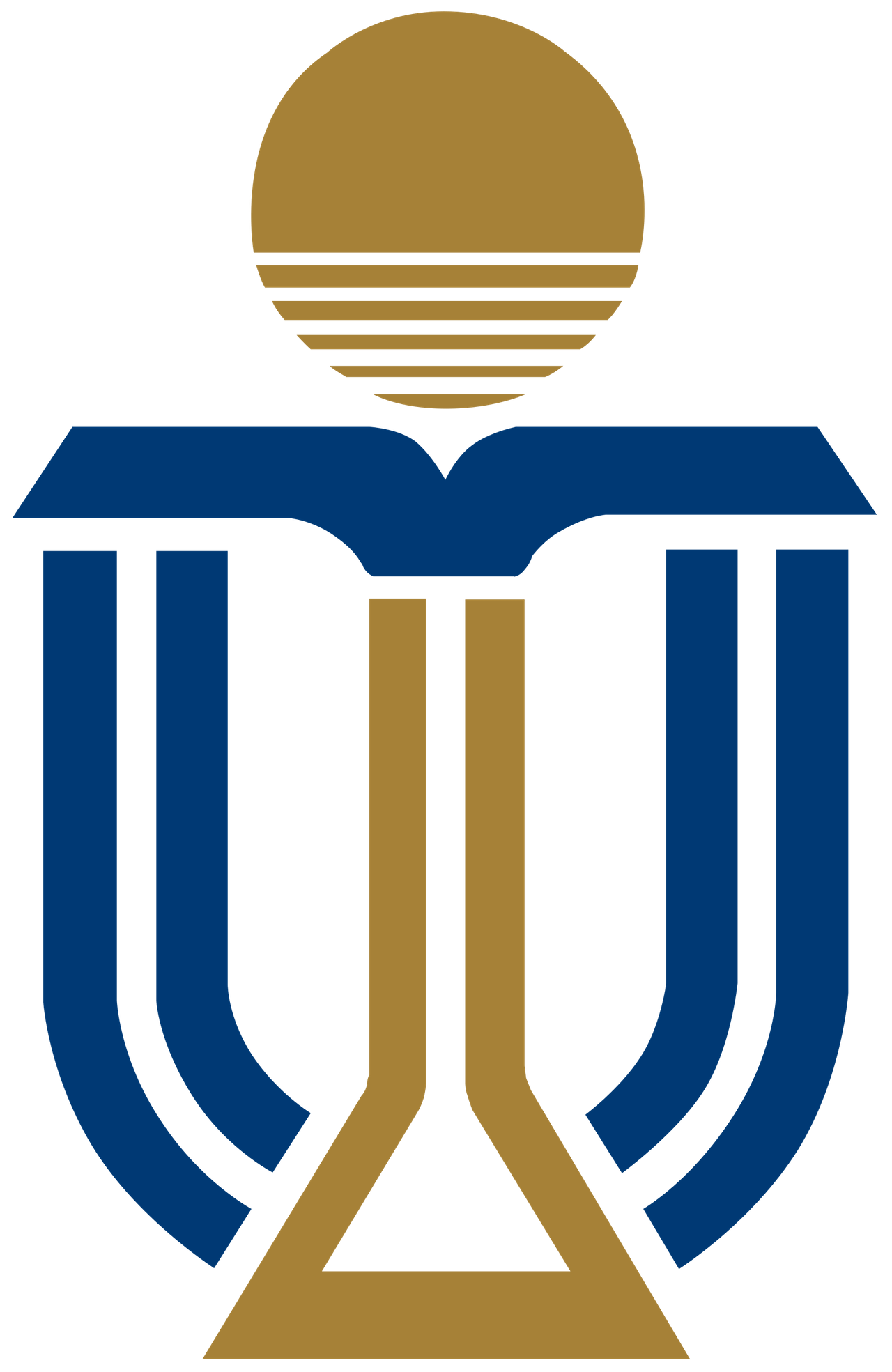}
\setabstractbrandlogoheight{1cm}

\usepackage{amsmath,amssymb,amsfonts}
\usepackage{mathtools}
\usepackage{amsthm}
\usepackage{bbm}
\usepackage{bm}
\usepackage{subfigure}
\usepackage{float}
\usepackage{adjustbox}
\usepackage{colortbl}
\usepackage{arydshln}
\usepackage{algorithm,algpseudocode}
\usepackage{algorithmicx}
\usepackage{wrapfig}
\usepackage{pifont}
\usepackage{bbding}
\usepackage{xspace}
\usepackage{multicol}
\usepackage{array}
\usepackage{multirow}
\usepackage{makecell}
\usepackage{pgf}
\usepackage{enumitem}
\usepackage{tabularx}
\usepackage{longtable}
\usepackage{xurl}
\usepackage{nicefrac}

\newcolumntype{Y}{>{\raggedright\arraybackslash}X}
\newcolumntype{P}[1]{>{\raggedright\arraybackslash}p{#1}}

\setlist[itemize]{leftmargin=1.2em,itemsep=1pt,topsep=2pt}
\setlist[enumerate]{leftmargin=1.4em,itemsep=1pt,topsep=2pt}
\definecolor{myblue}{HTML}{76C7E5}
\newcommand{\heatcell}[3]{%
    \pgfmathsetmacro{\percent}{(#1 - #2) / (#3 - #2) * 50 + 10}%
    \edef\temp{\noexpand\cellcolor{myblue!\percent}\relax \noexpand\makebox[2.2em][r]{#1}}%
    \temp
}
\newcommand{\heatcellb}[3]{%
    \pgfmathsetmacro{\percent}{(#1 - #2) / (#3 - #2) * 50 + 10}%
    \edef\temp{\noexpand\cellcolor{myblue!\percent}\relax \noexpand\makebox[2.2em][r]{\noexpand\textbf{#1}}}%
    \temp
}

\makeatletter
\DeclareRobustCommand\onedot{\futurelet\@let@token\@onedot}
\def\@onedot{\ifx\@let@token.\else.\null\fi\xspace}

\makeatother

\usepackage{graphicx}
\usepackage{xcolor}
\usepackage{amsmath,amssymb,amsfonts}
\usepackage{mathtools}
\usepackage{bm}
\usepackage{bbm}
\usepackage{booktabs}
\usepackage{float}
\usepackage{adjustbox}
\usepackage{colortbl}
\usepackage{xspace}
\usepackage{array}
\usepackage{multirow}
\usepackage{makecell}
\usepackage{tabularx}
\usepackage{longtable}
\usepackage{pifont}
\usepackage{tikz}
\usetikzlibrary{arrows.meta,calc,decorations.text,fit,positioning,shapes.geometric}
\usepackage{hyperref}
\usepackage{xurl}

\definecolor{pathblue}{HTML}{176B87}
\definecolor{pathgreen}{HTML}{3B7A57}
\definecolor{pathred}{HTML}{B6413A}
\definecolor{pathgold}{HTML}{B57918}
\definecolor{pathlight}{HTML}{EDF4F6}
\definecolor{pathhema}{HTML}{5A4778}
\definecolor{pathhemawash}{HTML}{EEEAF4}
\definecolor{patheosinwash}{HTML}{F8E7ED}

\newcolumntype{Y}{>{\raggedright\arraybackslash}X}
\newcolumntype{P}[1]{>{\raggedright\arraybackslash}p{#1}}
\newcommand{\cmark}{\textcolor{pathgreen}{\ding{51}}}
\newcommand{\xmark}{\textcolor{pathred}{\ding{55}}}
\newcommand{\partialmark}{\textcolor{pathgold}{\textbf{P}}}
\newcommand{\ours}{\textsc{PathoArgus-Bench}\xspace}
\newcommand{\pathoargus}{\textsc{PathoArgus}\xspace}
\newcommand{\figref}[1]{Fig.~\ref{#1}}
\newcommand{\tabref}[1]{Tab.~\ref{#1}}
\newcommand{\secref}[1]{Sec.~\ref{#1}}
\newcommand{\finding}[1]{\medskip\noindent\textbf{Finding #1.}}

\title{\pathoargus: Advancing Evidence-Grounded Long-Context Visual Reasoning across Gigapixel Whole-Slide and Multi-Slide Case Contexts}

\author[1]{Bowen Liu}
\author[1]{Qixiang Zhang}
\author[1]{Xiaomeng Li}
\affiliation[1]{The Hong Kong University of Science and Technology}

\correspondence{Xiaomeng Li}

\abstract{%
Whole-slide pathology reasoning requires models to integrate gigapixel-scale visual evidence across complete case-linked slides, yet current question-answering benchmarks primarily measure final answer accuracy—a metric vulnerable to linguistic priors and benchmark regularities, and insufficient to establish that predictions are grounded in the supplied tissue. We introduce \ours, a benchmark and evaluation protocol that explicitly tests the full evidence chain: availability, accessibility, use, and responsiveness. \ours{} comprises 22,078 four-choice questions from 4,913 patients across 15 TCGA projects, covering six pathology capabilities across three levels of evidence demand, and operates under a fixed reader budget that retains only a small fraction of the gigapixel context. To further isolate evidence-grounded reasoning, we contribute ESG (Evidence State Quartets), a controlled set of 483 quartets where the question text is fixed while the target WSI set is moved, replaced, or removed, requiring consistent predictions across all states. Evaluating 20 general-purpose, medical, and pathology-specific systems reveals a stark gap: while GPT-5.6 achieves 57.09\% overall accuracy and 57.04\% on ESG, it correctly completes only 19 of 483 quartets (3.93\% QExact), exposing that row-level accuracy does not translate into reliable evidence grounding. We also introduce \pathoargus, a fixed-budget reader that allocates context via question relevance and spatial coverage, attaining 50.39\% overall accuracy yet only 1.86\% QExact—demonstrating that improved context access alone does not ensure consistent evidence-based prediction. Our benchmark and diagnostics establish that acquiring useful whole-slide context is necessary but far from sufficient, and call for a shift from answer-centric to evidence-grounded evaluation in computational pathology.
}
\metadata[Data]{\url{https://huggingface.co/datasets/liubw/PathoArgus-Bench}}

\begin{document}

\maketitle

\input{Introduction}
\input{Related_work}

\input{Benchmark}
\input{Method}
\input{exp}
\input{Conclusion}

\bibliographystyle{assets/plainnat}
\bibliography{reference}

\end{document}

%% file: Introduction.tex
\section{Introduction}
\label{sec:introduction}

Whole-slide pathology reasoning asks a model to turn complete, case-linked
collections of gigapixel slides into pathology decisions. Unlike recognition
from a prepared crop, this setting couples visual reasoning with evidence
acquisition: a reader must recover question-relevant tissue from tens of
thousands of patches, preserve the appropriate spatial scale, integrate
information across slides, and reject irrelevant case context. Reliable
performance therefore requires an evidence chain. The target tissue must be
present in the supplied case, survive a limited reader budget, influence the
answer, and induce the correct response when the evidence changes.

\begin{figure*}[t]
\centering
\includegraphics[width=\textwidth]{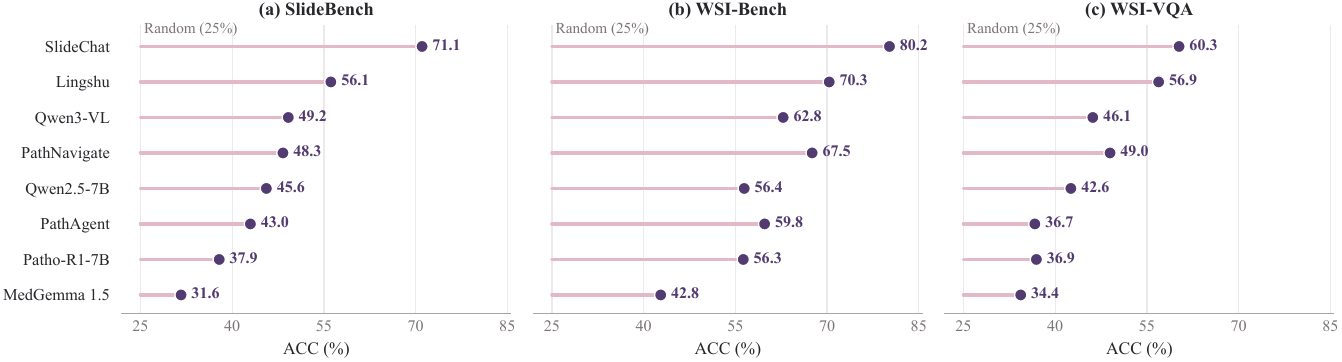}
\caption{Text-only ACC remains high on SlideBench, WSI-Bench and WSI-VQA. Each model receives only the question and answer choices.}
\label{fig:text-only-motivation}
\end{figure*}

Conventional question-answering accuracy measures only the endpoint of this
chain. MIRAGE shows that MLLMs can retain substantial medical and general VQA
performance without image input by exploiting question structure, answer
priors, and benchmark regularities~\citep{asadi2026mirage}. The same issue is
visible in pathology. Our no-image audit gives eight baselines only the question
and answer choices from SlideBench~\citep{chen2025slidechat},
WSI-Bench~\citep{liang2025wsillava} and WSI-VQA~\cite{chen2024wsivqa}. As
\figref{fig:text-only-motivation} shows, the strongest text-only ACC reaches
71.06\% on SlideBench and 80.20\% on the four-choice portion of WSI-Bench. On
SlideBench, this is 29.69 points above the 41.37\% majority-position baseline.
Answer accuracy alone therefore cannot establish that a model used WSI
evidence.

Whole-slide input introduces an additional bottleneck that a no-image control
cannot diagnose. Even when every slide is supplied, answer-critical tissue may
be discarded before reaching the language model because only a small visual
budget can be retained. Evidence availability, accessibility, use, and
responsiveness are thus distinct. The last distinction becomes visible when a
question and its choices remain fixed while the target case moves among
candidate WSI sets or disappears. Scoring these conditions independently can
reward a coincidental answer; scoring the complete sequence tests whether
predictions follow the supplied evidence.

Existing pathology benchmarks cover important parts of this chain in separate
settings. Slide-level QA evaluates answer generation over bounded or encoded
visual inputs; navigation benchmarks study evidence seeking through specialized
interfaces; and grounding benchmarks probe visual dependence
~\citep{chen2024wsivqa,chen2025slidechat,liang2025wsillava,lu2026ctisqa,
buckley2025giant,liao2026pathagentbench,zhang2026pathbind,
chen2026pathviewbench}. What remains missing is a shared protocol for one
central question: \textbf{Can current MLLMs ground pathology decisions in
visual evidence from complete gigapixel slides and multi-slide cases?} \ours
operationalizes this question along three observable dimensions: performance
across six pathology capabilities, evidence acquisition under an explicit
reader budget, and prediction consistency under controlled changes to the
supplied WSI set.

\begin{figure*}[t]
\centering
\begin{minipage}[t]{0.62\textwidth}
\vspace{0pt}\centering
{\scriptsize\bfseries (a) Capability and evidence hierarchy\par}
\vspace{-1pt}
\includegraphics[width=\linewidth]{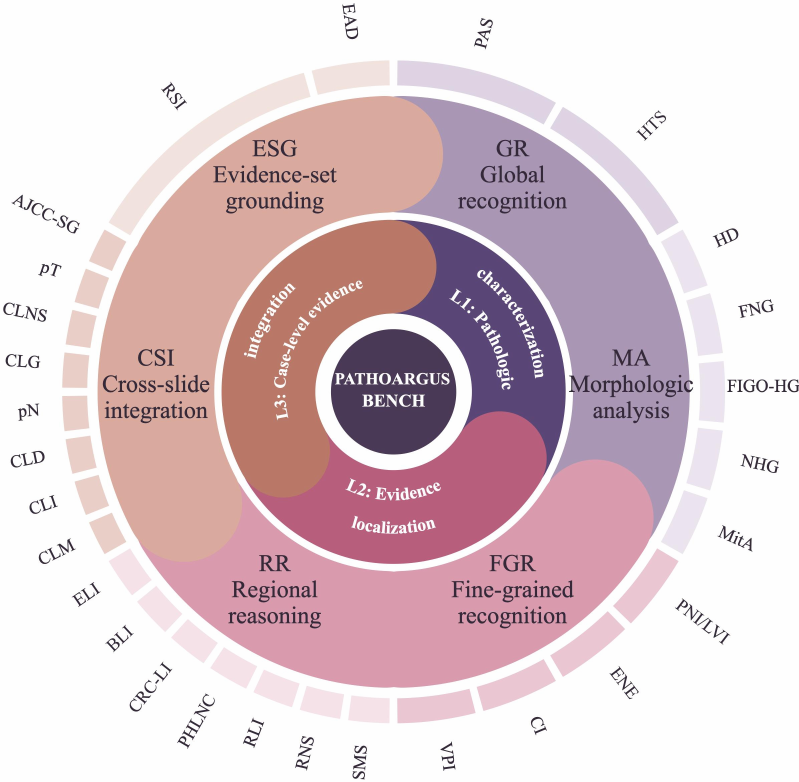}
\end{minipage}%
\begin{minipage}[t]{0.38\textwidth}
\vspace{0pt}\centering
{\scriptsize\bfseries (b) TCGA project coverage\par}
\vspace{-1pt}
\includegraphics[width=\linewidth]{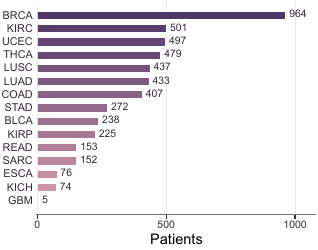}

\vspace{-2pt}
{\scriptsize\bfseries (c) Long-context workload\par}
\vspace{-1pt}
\includegraphics[width=\linewidth]{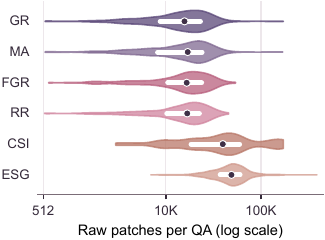}
\end{minipage}
\caption{Overview of \ours: (a) capability hierarchy, (b) TCGA project
coverage, and (c) long-context patch workload. Panel (c) shows medians (dots)
and interquartile ranges (white bars) on a log scale for questions with at
least 512 patches.}
\label{fig:benchmark-overview}
\end{figure*}

We introduce \ours to instantiate this evaluation target. Every model begins
from complete case-linked WSI context and operates under an explicit reader
budget. Six tasks form three levels of evidence demand: pathologic
characterization, localized evidence assessment, and case-level evidence
integration (\figref{fig:benchmark-overview}(a)). The benchmark contains 22,078
four-choice questions from 4,913 patients across 15 TCGA projects, partitioned
into 15,702 training, 2,095 validation, and 4,281 bench questions. A bench
question exposes 33,743 patch features on average and up to 388,637; a
$K=512$ reader retains 1.51\% of the aggregate context. ESG adds 483 controlled
quartets that hold the text fixed while moving or removing the target WSI set,
and QExact credits a quartet only when all four evidence states are answered
correctly.

The evaluation exposes two gaps. First, model rankings conceal sharply uneven
capability profiles. Second, row-level accuracy does not translate into
consistent evidence-grounded reasoning: GPT-5.6 reaches 57.09\% Overall and
57.04\% ESG accuracy, yet completes only 19 of 483 quartets (3.93\% QExact).
We further introduce \pathoargus, a fixed-budget companion reader that targets
the accessibility stage by allocating context according to question relevance,
candidate-set structure, and slide-spatial coverage. At $K=512$, it reaches
50.39\% Overall and 46.17\% ESG accuracy, while its 1.86\% QExact shows that
improved context allocation does not by itself establish consistent evidence
use. The benchmark and method therefore support one conclusion: acquiring
useful WSI context is necessary, but evidence-grounded prediction remains a
separate challenge.

Our main contributions are summarized as follows:
\begin{itemize}
    \item We formulate complete-context WSI question answering as an evidence
    chain and introduce \ours, which evaluates six pathology capabilities under
    explicit gigapixel-to-multi-slide context and reader budgets.
    \item We evaluate 20 general-purpose, medical, and pathology-specific
    systems using row-level, capability-level, text-only, and controlled-quartet
    diagnostics. The results separate capability breadth, evidence access, and
    evidence-grounded prediction.
    \item We introduce \pathoargus, a fixed-budget reader that allocates context
    by question relevance, candidate-set structure, and slide-spatial coverage,
    providing a concrete operating point for the evidence-accessibility stage.
\end{itemize}

%% file: Related_work.tex
\section{Related Work}
\label{sec:related-work}

\paragraph{Whole-slide pathology question answering.}
PathVQA introduced visual question answering over conventional pathology
images~\citep{he2021pathvqa}, while PathMMU broadened single-image evaluation
with pathologist-reviewed multiple-choice questions and text-only shortcut
controls~\citep{sun2024pathmmu}. Broader medical suites such as OmniMedVQA and
GMAI-MMBench include histopathology among diverse image modalities
~\citep{hu2024omnimedvqa,chen2024gmaimmb}, while MicroVQA evaluates expert
reasoning over biological microscopy with no-image controls
~\citep{burgess2025microvqa}.
WSI-VQA reframed slide-level pathology tasks as generative question answering
over 8,672 pairs and 977 WSIs~\citep{chen2024wsivqa}. SlideChat introduced
SlideBench for WSI captioning and question answering across microscopy,
diagnosis, and clinical settings~\citep{chen2025slidechat}. WSI-LLaVA developed
the morphology-aware WSI-Bench with approximately 180K question-answer pairs
and four pathology capability groups~\citep{liang2025wsillava}. These
benchmarks establish broad pathology QA coverage. Their scores primarily
characterize answer quality after the visual input has been bounded or encoded;
\ours explicitly evaluates both the complete case-linked context and the subset
retained by a reader.

\input{tables/benchmark_comparison}

\paragraph{Evidence acquisition and multiscale reasoning.}
GIANT iteratively navigates WSIs and accompanies MultiPathQA, which includes
five clinically relevant question types~\citep{buckley2025giant}. PathNavigate
combines a surprise-guided
global scan, shared slide memory, and question-conditioned local search
~\citep{yang2026pathnavigate}, while PathAgent organizes WSI analysis as
iterative navigation, perception, and evidence integration
~\citep{chen2026pathagent}. These methods make evidence search explicit
through distinct interfaces, visual budgets, and task scopes. \ours places
dense, regional, sparse, cross-slide, and multi-set evidence under one
fixed-budget protocol. Within this protocol, \pathoargus combines relevance and
coverage across candidate sets, slides, and spatial regions.

%% file: tables/benchmark_comparison.tex
\begin{table}[t]
\centering
\small
\caption{Coverage of the evaluation chain from whole-slide context to
evidence-grounded reasoning; ``P'' denotes partial coverage. The final axis is
operationalized by holding the question fixed while changing the target WSI
set.}
\label{tab:benchmark-comparison}
\resizebox{\textwidth}{!}{%
\begin{tabular}{lccccc}
\toprule
Benchmark & Whole-slide context & Evidence acquisition & Visual dependence & Multi-slide reasoning & Evidence-grounded reasoning \\
\midrule
PathVQA~\citep{he2021pathvqa} & \xmark & \xmark & \xmark & \xmark & \xmark \\
OmniMedVQA~\citep{hu2024omnimedvqa} & \xmark & \xmark & \xmark & \xmark & \xmark \\
GMAI-MMBench~\citep{chen2024gmaimmb} & \xmark & \xmark & \xmark & \xmark & \xmark \\
PathMMU~\citep{sun2024pathmmu} & \xmark & \xmark & \cmark & \xmark & \xmark \\
WSI-VQA~\citep{chen2024wsivqa} & \cmark & \xmark & \xmark & \xmark & \xmark \\
MicroVQA~\citep{burgess2025microvqa} & \xmark & \xmark & \cmark & \xmark & \xmark \\
SlideBench~\citep{chen2025slidechat} & \cmark & \xmark & \partialmark & \xmark & \xmark \\
WSI-Bench~\citep{liang2025wsillava} & \cmark & \xmark & \xmark & \xmark & \xmark \\
\rowcolor{pathhemawash}\textbf{PathoArgus-Bench} & \cmark & \cmark & \cmark & \cmark & \cmark \\
\bottomrule
\end{tabular}}
\end{table}

%% file: Benchmark.tex
\section{PathoArgus-Bench}
\label{sec:benchmark}

\subsection{Evaluation Target}

Given a question $q$, four semantic choices
$\mathcal{A}=\{a_1,\ldots,a_4\}$, and a case-linked collection of patch
features $\mathcal{X}=\{x_1,\ldots,x_N\}$ from one or more WSIs, a system reads
at most $K$ features and predicts one choice. \ours records both the
available context $N$ and reader context $K$, making context compression part
of the evaluation protocol. ESG generalizes the input to three candidate WSI
sets and adds an evidence-absence choice for questions whose target tissue is
not present.

The benchmark distinguishes four stages of evidence-grounded reasoning.
Evidence is \emph{available} when target-relevant tissue occurs in the supplied
case context and \emph{accessible} when it survives the $K$-feature reader
budget. It is \emph{used} when the prediction depends on visual rather than
textual cues and \emph{responsive} when a controlled change in evidence induces
the corresponding answer change. The recorded $N$ and $K$ characterize the
availability and access constraints. Text-only evaluation probes non-visual
inference, while ESG evidence-set interventions probe responsiveness.

Three design principles operationalize these stages. First, models begin from
complete case-linked WSI context rather than an answer-specific crop. Second,
dense, regional, sparse, multi-slide, and multi-set evidence share the same
budgeted interface. Third, patient- and WSI-disjoint splits test case-level
generalization, while all four ESG interventions on a semantic question remain
grouped during evaluation.

\subsection{Capability and Evidence-Organization Taxonomy}

\input{tables/task_taxonomy}

We abbreviate the six capabilities as GR (Global Recognition), MA
(Morphologic Analysis), FGR (Fine-Grained Recognition), RR (Regional
Reasoning), CSI (Cross-Slide Integration), and ESG (Evidence-Set Grounding).
\tabref{tab:taxonomy} organizes the benchmark as a progression of evidence
demands. GR and MA characterize what a case shows through global recognition
and morphologic analysis. FGR and RR require localized assessment of
fine-grained or regional findings. CSI integrates evidence across multiple
slides. ESG compares three candidate WSI sets and determines either which set contains the target or
that target evidence is absent. Its 483 semantic questions each instantiate
four evidence states, yielding 1,932 condition-specific rows.

For readability, \figref{fig:benchmark-overview}(a) uses initialisms for the
outer-ring subtasks. PAS and HTS denote primary anatomic site and histologic
type/subtype. HD, FNG, FIGO-HG, NHG, and MitA denote histologic differentiation,
Fuhrman nuclear grade, FIGO histologic grade, Nottingham histologic grade, and
mitotic activity. PNI/LVI denotes perineural or lymphovascular invasion, while
ENE, CI, and VPI denote extranodal extension, capsular invasion, and visceral
pleural invasion. SMS and RNS denote surgical margin status and regional nodal
status; RLI, CRC-LI, BLI, and ELI denote renal, colorectal, bladder, and
endometrial local invasion; and PHLNC denotes positive H\&E lymph node count.
CLM, CLI, CLD, CLG, and CLNS denote case-level margin, invasion, diagnosis,
grade, and nodal status. The standard $pT$ and $pN$ labels denote pathologic T
and N categories. AJCC-SG denotes AJCC stage group, while RSI and EAD denote
relevant-set identification and evidence-absence detection.

\subsection{Construction Pipeline}

\ours combines reverse synthesis for capability coverage with controlled
injection for evidence responsiveness. We begin with 4,962 patients and 5,516
locally registered TCGA WSIs~\citep{tcga2013pancancer}. For the five single-set
capabilities (GR, MA, FGR, RR, and CSI), structured
pathology report fields covering anatomic site, diagnosis, grade, invasion,
margin, nodal status, and stage are mapped to intent-specific questions and
four-choice semantic families, followed by answer-position permutation.

FGR preserves rare joint perineural and lymphovascular invasion labels through
class-aware sampling and training-time query augmentation. For ESG, controlled
injection places the target case in candidate WSI Set 1, 2, or 3, or removes it
from all sets. Donor cases come from the same split and a different anatomic
site, with their WSI counts matched to the target case. The question, choices,
and candidate-set capacities remain fixed across the quartet; only the target's
location or presence changes across its four conditions.

We assign 4,913 patients using project-, FGR-label-, and WSI-count-aware
stratification. The runtime projection removes answers, case identities,
counterfactual conditions, decision metadata, and construction fields.
Automated quality checks verify patient isolation, WSI isolation, metadata
separation, answer-position balance, and ESG donor constraints.

\subsection{Scale, Splits, and Context Workload}

\input{tables/dataset_statistics}

\ours contains 22,078 questions over 4,809 target cases and 5,400 WSIs. The
training, validation, and bench splits contain 15,702, 2,095, and 4,281
questions, respectively. The bench split covers 959 target cases, with 977
input cases and 1,099 unique WSIs after incorporating ESG donors. Each TCGA
project is within one patient of its target 70/10/20 allocation, with zero
patient or WSI overlap across splits.

\input{tables/context_statistics}

The complete-context protocol exposes substantial variation in visual workload.
As shown in \figref{fig:benchmark-overview}(c) and \tabref{tab:context}, 85.52\% of bench
questions contain more than 10,000 patch features, 22.70\% contain more than
50,000, and the largest contains 388,637. Under a $K=512$ reader budget, the benchmark
contains 144,453,816 available features and 2,183,466 selected features,
corresponding to an aggregate retention rate of 1.51\%. The benchmark therefore
tests evidence acquisition rather than assuming that answer-critical tissue has
already been retained.

\subsection{Non-Visual and Structural Controls}

Answer positions are balanced to within one question in every split and within
two questions in each bench capability. A Qwen2.5-7B text-only control obtains
24.39\% Overall accuracy and 0/483 QExact; 96.07\% of its ESG quartets receive a
constant prediction across all evidence states. Its 30.87\% FGR accuracy shows
that aggregate answer-position balance does not remove capability-specific
textual signal. We therefore report capability-level performance and audit
semantic choice support in
Appendix~\ref{sec:construction-audits}, separating answer-position balance from
the empirical support of individual clinical options.

%% file: tables/task_taxonomy.tex
\begin{table}[t]
\centering
\small
\caption{Three-level capability and evidence-organization taxonomy in the
bench split. ESG counts condition rows and groups.}
\label{tab:taxonomy}
\begin{tabularx}{\textwidth}{clYYr}
\toprule
Level & Task & Capability & Evidence organization & $N$ \\
\midrule
L1 & GR & Global recognition & slide/case; dense & 804 \\
L1 & MA & Morphologic analysis & region; medium & 429 \\
L2 & FGR & Fine-grained recognition & patch/region; sparse & 149 \\
L2 & RR & Regional reasoning & multi-region intent & 820 \\
L3 & CSI & Cross-slide integration & multi-slide input & 147 \\
L3 & ESG & Evidence-set grounding & multi-set counterfactual & 1,932 / 483 \\
\bottomrule
\end{tabularx}
\end{table}

%% file: tables/dataset_statistics.tex
\begin{table}[t]
\centering
\small
\caption{Patient-isolated split statistics. ``Input cases'' includes ESG donor
cases; target cases produce at least one target QA.}
\label{tab:dataset-stats}
\resizebox{\textwidth}{!}{%
\begin{tabular}{lrrrrrr}
\toprule
Split & Assigned patients & Target cases & Input cases & Unique WSIs & QA rows & Share \\
\midrule
Training & 3,439 & 3,372 & 3,424 & 3,768 & 15,702 & 71.12\% \\
Validation & 491 & 478 & 484 & 533 & 2,095 & 9.49\% \\
Bench & 983 & 959 & 977 & 1,099 & 4,281 & 19.39\% \\
\midrule
Total & 4,913 & 4,809 & 4,885 & 5,400 & 22,078 & 100\% \\
\bottomrule
\end{tabular}}
\end{table}

%% file: tables/context_statistics.tex
\begin{table}[t]
\centering
\small
\caption{Available patch-feature workload in the benchmark split.}
\label{tab:context}
\begin{tabular}{lr@{\qquad}lr}
\toprule
Statistic & Patches & Context bucket & Rows (share) \\
\midrule
Mean & 33,743 & $\leq 512$ & 21 (0.49\%) \\
Median & 26,931 & 513--2,000 & 78 (1.82\%) \\
P90 & 63,694 & 2,001--10,000 & 521 (12.17\%) \\
P95 & 73,467 & 10,001--50,000 & 2,689 (62.81\%) \\
P99 & 140,037 & $>50,000$ & 972 (22.70\%) \\
Maximum & 388,637 & Aggregate K512 retention & 1.51\% \\
\bottomrule
\end{tabular}
\end{table}

%% file: Method.tex
\section{PathoArgus}
\label{sec:method}

\subsection{Overview}

\pathoargus addresses the evidence-access bottleneck created when complete WSI
context must be compressed into a limited reader budget. A conventional
selector flattens all patches into one list and retains those with the highest
question relevance. This loses the structure of the case: a large candidate
set can suppress a smaller one, one slide can dominate a multi-slide case, and
high-scoring patches can cluster within a narrow tissue region. Once such
evidence is removed, the downstream reader cannot recover it.

Our central idea is to preserve the units that must be compared before ranking
their patches. PathoArgus treats evidence selection as routing through the
hierarchy
\emph{candidate set $\rightarrow$ slide $\rightarrow$ spatial region
$\rightarrow$ patch}. It first keeps every supplied candidate context
accessible, then distributes the visual budget across the case hierarchy, and
finally selects patches by combining question relevance with spatial coverage.
The result is a compact, ordered visual sequence for the WSI reader.

\subsection{Structure-Preserving Budget Routing}

The first challenge is structural imbalance. In an ordinary question, the
visual input forms one case context; in ESG, the input contains several
candidate contexts that must remain comparable. Applying a global candidate
limit or Top-$K$ operation can erase an entire context before the reader sees
it, turning a comparison problem into a guess over incomplete evidence.

PathoArgus therefore preserves candidate-set identity throughout selection.
Before relevance ranking, it constructs a candidate pool in which every
supplied set remains represented. It then allocates the reader budget across
sets subject to their available content: a single-set case receives the full
budget, while a multi-set case reserves capacity for each comparison context.
Unused capacity from a small set is reassigned rather than discarded.

The same routing principle is applied within each set. For cases containing
multiple slides, PathoArgus reserves coverage across slides before choosing
individual patches. This hierarchical allocation separates two decisions that
global ranking conflates: \emph{where evidence must remain accessible} and
\emph{which evidence is most relevant within that context}.

\subsection{Query-Aware Relevance-Coverage Selection}

Set and slide allocation still leaves a spatial selection problem. Pure
relevance ranking can repeatedly select nearby patches from the same tissue
focus, whereas uniform sampling can spend much of the budget on regions that
do not address the question. PathoArgus couples relevance with spatial
coverage.

Within each routed context, a coverage branch spreads selections across slides
and occupied spatial regions. Question relevance determines the representative
chosen from each region, so coverage remains query aware. A relevance branch
then uses the remaining capacity for the highest-scoring unselected patches,
retaining concentrated evidence that broad coverage may miss.

Finally, the two branches are merged, duplicate patches are removed, and the
selected evidence is restored to its original WSI order before being passed to
the reader. Candidate-set structure determines comparability, slide-spatial
coverage preserves distributed morphology, and question relevance concentrates
the remaining budget on likely evidence. Together, these steps make evidence
access explicit while leaving answer generation to the downstream reader.

%% file: exp.tex
\section{Experiments}
\label{sec:experiments}

\subsection{Evaluation Protocol}

We evaluate on the 4,281-question bench split under a closed-set A--D answer
protocol. The broad comparison covers 20 pretrained general-purpose, medical,
and pathology systems. Standard image-based MLLMs receive a 1024-pixel WSI
overview and 16 deterministic views at both $5\times$ and $20\times$
magnification; SlideChat and WSI-LLaVA use their official native WSI feature
interfaces. We also adapt the question-conditioned traversal strategies of
PathNavigate~\citep{yang2026pathnavigate} and
PathAgent~\citep{chen2026pathagent} to the Qwen3-VL evaluation interface.
The supervised baseline is fine-tuned on the \ours training split.

\pathoargus uses CONCH patch features~\citep{lu2024conch}, a Qwen2.5-7B
reader~\citep{bai2025qwen25vl}, and a budget of $K=512$ selected patches, as
detailed in \secref{sec:method}. Its selector scores at most $M=10{,}000$
candidates per question.

We report Overall accuracy together with per-capability accuracy and QExact.
Overall measures row-level utility, while QExact tests whether all four
conditions in an ESG quartet are answered correctly:

\begin{equation}
\begin{aligned}
    \operatorname{Overall}
    &=\frac{1}{N}\sum_{i=1}^{N}\mathbbm{1}[\hat y_i=y_i],\\
    \operatorname{QExact}
    &=\frac{1}{G}\sum_{g=1}^{G}
      \mathbbm{1}\!\left[\bigwedge_{c=1}^{4}\hat y_{g,c}=y_{g,c}\right],
\end{aligned}
\label{eq:evaluation-metrics}
\end{equation}

where $N=4{,}281$ and $G=483$.

\subsection{Capability across Pathology Evidence Demands}

\input{tables/main_results}

\finding{1} \textbf{Strong aggregate performance does not imply broad
capability coverage.} GPT-5.6 achieves the highest Overall accuracy at 57.09\%,
whereas the strongest pretrained open model, Lingshu-32B, reaches 36.74\%.
The strongest evaluated pathology-specific model, WSI-LLaVA, obtains 30.46\%,
and the \ours-supervised baseline obtains 32.36\%. The companion \pathoargus
reader reaches 50.39\% under its 512-patch budget.

Capability profiles expose variation hidden by Overall accuracy. GPT-5.6
reaches 85.20\% on GR and 57.04\% on ESG but only 24.16\% on FGR. Specialized
systems show similarly localized strengths: WSI-LLaVA reaches 51.74\% on GR,
MedVLM-R1-2B reaches 42.68\% on RR, and PathAgent reaches 34.69\% on CSI. In
this evaluation, these peaks remain capability-specific rather than forming a
uniformly strong profile across all six tasks.

\subsection{Does the Answer Follow the Evidence?}

\finding{2} \textbf{Row accuracy and evidence responsiveness remain sharply
separated.} The text-only control reaches 24.95\% ESG accuracy but 0\% QExact,
with 96.07\% of quartets receiving one constant prediction across all four
evidence states. GPT-5.6 reaches 57.04\% ESG accuracy, yet completes only 19
of 483 quartets (3.93\%). High row accuracy therefore does not establish that
predictions track controlled changes in the supplied WSI evidence.

%% file: tables/main_results.tex
\begin{table}[t]
\centering
\small
\setlength{\tabcolsep}{3pt}
\renewcommand{\arraystretch}{1.02}

\label{tab:main-results}
\resizebox{\linewidth}{!}{%
\arrayrulecolor{pathhema}
\begin{tabular}{@{}lccccccccc@{}}
\toprule
\textbf{Model} & \textbf{Size} & \textbf{Overall}$\uparrow$
& \textbf{GR}$\uparrow$ & \textbf{MA}$\uparrow$ & \textbf{FGR}$\uparrow$
& \textbf{RR}$\uparrow$ & \textbf{CSI}$\uparrow$ & \textbf{ESG}$\uparrow$
& \textbf{QExact}$\uparrow$ \\
\midrule
\rowcolor{patheosinwash}
\multicolumn{10}{l}{\textcolor{pathhema}{$\blacktriangledown$ \emph{General-purpose MLLMs}}} \\
GPT-5.6~\citep{openai2026gpt56} & -- & \textbf{57.09} & \textbf{85.20} & \underline{46.85} & 24.16
& \underline{42.93} & \textbf{46.26} & \textbf{57.04} & \textbf{3.93} \\
InternVL3~\citep{zhu2025internvl3} & 8B & 27.12 & 37.44 & 16.32 & \textbf{37.58} & 22.80 & 26.53
& 26.29 & 0.00 \\
Qwen3-VL~\citep{bai2025qwen3vl} & 8B & 26.47 & 36.19 & 23.54 & 18.12 & 22.44 & 31.29 & 25.05 & 0.00 \\
Qwen3-VL-A3B~\citep{bai2025qwen3vl} & 30B-A3B & 26.42 & 33.96 & 25.41 & 24.83 & 24.63 & 25.85 & 24.43 & 0.00 \\
GLM-4.6V-Flash~\citep{vteam2025glm45v} & 10B & 24.69 & 25.00 & 23.08 & 24.83 & 25.00 & 23.81 & 24.84 & 0.00 \\
Vision-DeepResearch-8B~\citep{huang2026visiondeepresearch} & 8B & 24.67 & 26.87 & 17.72 & 23.49 & 24.39 & 29.93 & 25.10 & 0.00 \\
\midrule
\rowcolor{patheosinwash}
\multicolumn{10}{l}{\textcolor{pathhema}{$\blacktriangledown$ \emph{Medical MLLMs}}} \\
LLaVA-Med v1.5~\citep{li2023llavamed} & 7B & 24.83 & 25.00 & 24.94 & 24.83 & 24.27 & 24.49 & 25.00 & 0.00 \\
MedVLM-R1-2B~\citep{pan2025medvlmr1} & 2B & 29.60 & 35.07 & 16.55 & 28.86 & 42.68 & 27.89 & 24.84 & 0.00 \\
HuatuoGPT-V~\citep{chen2024huatuogptvision} & 7B & 25.67 & 38.68 & 16.55 & 20.13 & 19.51 & 27.21 & 25.21 & 0.00 \\
HuatuoGPT-V~\citep{chen2024huatuogptvision} & 34B & 27.54 & 39.30 & 42.42 & 26.85 & 15.24 & 28.57 & 24.53 & 1.24 \\
Lingshu~\citep{xu2025lingshu} & 7B & 35.76 & 64.93 & 41.72 & 33.56 & 27.93 & 29.93 & 26.24 & 0.21 \\
Lingshu~\citep{xu2025lingshu} & 32B & 36.74 & 66.04 & \textbf{47.32}
& \underline{36.24} & 25.24 & \underline{32.65} & 27.43 & 0.83 \\
MedGemma 1.5~\citep{sellergren2026medgemma15} & 4B & 26.14 & 36.07 & 15.38 & 26.85 & 23.66 & 29.25 & 25.16 & 0.00 \\
MedGemma~\citep{sellergren2025medgemma} & 27B & 29.01 & 33.21 & 28.21 & \textbf{37.58} & 33.05
& 31.97 & 24.84 & 0.00 \\
\midrule
\rowcolor{patheosinwash}
\multicolumn{10}{l}{\textcolor{pathhema}{$\blacktriangledown$ \emph{Pathology-specific methods}}} \\
Quilt-LLaVA~\citep{seyfioglu2024quiltllava} & 7B & 25.20 & 25.12 & 24.01 & 24.83 & 26.71 & 23.81 & 25.00 & 0.00 \\
Patho-R1-7B~\citep{zhang2025pathor1} & 7B & 28.85 & 56.59 & 13.29 & 29.53 & 19.39 & 25.17 & 25.00 & 0.00 \\
SlideChat~\citep{chen2025slidechat} & $\sim$7B & 25.91 & 25.25 & 23.08 & 28.86 & 26.83 & 27.21 & 26.09 & 0.00 \\
WSI-LLaVA~\citep{liang2025wsillava} & $\sim$7B & 30.46 & 51.74 & 25.17 & 24.16 & 26.22 & 29.93 & 25.10 & 0.00 \\
PathNavigate~\citep{yang2026pathnavigate} & -- & 25.91 & 32.96 & 21.21 & 20.13 & 23.05 & 31.29 & 25.26 & 0.21 \\
PathAgent~\citep{chen2026pathagent} & -- & 25.46 & 32.84 & 20.28 & 19.46 & 23.05 & 34.69 & 24.33 & 0.00 \\
\midrule
\midrule
\rowcolor{patheosinwash}
\multicolumn{10}{l}{\textcolor{pathhema}{$\blacktriangledown$ \emph{PathoArgus}}} \\
SFT & 7B & 32.36 & 54.73 & 29.28 & 35.94 & 31.23 & 23.29 & 24.90 & 0.00 \\
\rowcolor{pathhemawash}
\textcolor{pathhema}{\pathoargus} & 7B & \underline{50.39} & \underline{72.89} & 36.36 & 32.89
& \textbf{49.51} & \textbf{46.26} & \underline{46.17} & \underline{1.86} \\
\bottomrule
\end{tabular}%
\arrayrulecolor{black}
}
\caption{Results on \ours. Bold and underline
denote the best and second-best results, respectively.}
\end{table}

%% file: Conclusion.tex
\section{Conclusion}

\ours asks whether current MLLMs can ground pathology decisions in visual
evidence from complete gigapixel slides and multi-slide cases. It operationalizes
this question through six pathology capabilities, explicit context accounting,
and controlled evidence-set quartets. The results provide a qualified answer:
across 20 baseline systems, GPT-5.6 reaches 57.09\% Overall accuracy, yet the
best QExact remains 3.93\%. The companion \pathoargus reader combines
question-conditioned relevance with candidate-set and slide-spatial coverage,
reaching 50.39\% Overall accuracy and 46.17\% ESG accuracy. Its 1.86\% QExact
shows that improved evidence access does not establish consistent
evidence-conditioned prediction. Together, these results identify evidence
access and evidence-responsive training as complementary directions for
long-context pathology reasoning.